# KeyXtract Twitter Model - An Essential Keywords Extraction Model for Twitter Designed using NLP Tools

Tharindu Weerasooriya[1#], Nandula Perera[2], S.R. Liyanage[3]

[1#]*Department of Statistics and Computer Science, University of Kelaniya, Sri Lanka*
[2]*Department of English, University of Kelaniya, Sri Lanka*
[3]*Department of Software Engineering, University of Kelaniya, Sri Lanka*
[1#]<cyriltcw@gmail.com>

***Abstract***— *Since a tweet is limited to 140 characters, it is ambiguous and difficult for traditional Natural Language Processing (NLP) tools to analyse. This research presents KeyXtract which enhances the machine learning based Stanford CoreNLP Part-of-Speech (POS) tagger with the Twitter model to extract essential keywords from a tweet. The system was developed using rule-based parsers and two corpora. The data for the research was obtained from a Twitter profile of a telecommunication company. The system development consisted of two stages. At the initial stage, a domain specific corpus was compiled after analysing the tweets. The POS tagger extracted the Noun Phrases and Verb Phrases while the parsers removed noise and extracted any other keywords missed by the POS tagger. The system was evaluated using the Turing Test. After it was tested and compared against Stanford CoreNLP, the second stage of the system was developed addressing the shortcomings of the first stage. It was enhanced using Named Entity Recognition and Lemmatization. The second stage was also tested using the Turing test and its pass rate increased from 50.00% to 83.33%. The performance of the final system output was measured using the $F_1$ score. Stanford CoreNLP with the Twitter model had an average $F_1$ of 0.69 while the improved system had a $F_1$ of 0.77. The accuracy of the system could be improved by using a complete domain specific corpus. Since the system used linguistic features of a sentence, it could be applied to other NLP tools.*

***Keywords*** *— **Natural Language Processing, Stanford CoreNLP, Tweet Analysis, Named Entity Recognition, Lemmatization, Keyword Extraction, Turing Test***

## I. INTRODUCTION

Natural Language Processing (NLP) has seen unprecedented development over the past two decades (Zitouni, 2014). Keyword extraction of NLP is used during Question and Answering (Q&A) processes.

In understanding a question, humans extract keywords that are vital in synthesizing the answer. These specific words can also be used to back-formulate the question. In NLP, POS tags could be used to extract key ideas from a sentence.

One of the most fertile grounds to put NLP to test is Twitter. A tweet might be ambiguous and is not always grammatically correct. Hence, conventional POS tagging methods cannot be used to extract keywords from a tweet.

Corporate giants often answer customer support requests through Twitter™, which has 320 million active users per month (*Twitter Usage / Company Facts*, 2016). In Sri Lanka, Dialog Axiata is a prominent telecommunication company that provides this service. Automating this process is challenging for a machine, as interpreting a tweet could be problematic.

This research presents KeyXtract which is a new utilization of the Stanford CoreNLP (Manning *et al.*, 2014) tool, a widely used machine learning based NLP tool. The research was conducted in two stages. The Twitter Model for KeyXtract presented in this paper is the extension of Stage 1 developed at the first stage of the research. In the first stage (Weerasooriya, Perera and S R Liyanage, 2016), Stanford CoreNLP was enhanced using parsers (to extract essential keywords using the linguistic features of a sentence) and a domain specific corpus (consisting of 206 words). The second stage presented in this paper consists of improvements made based on the evaluation results of stage 1. The Turing test was used to evaluate the success of this method in imitating the human logic, and its performance was measured using the $F_1$ score.

## II. RELATED WORK

### A. Extracting keywords

Mitkov and Ha, state that to extract a " 'keyword phrase', a list of semantically close terms including a *noun phrase, verb phrase, adjective phrase and adverb phrase*" (Mitkov and Ha, 1999) should be considered. In the current study, Noun Phrase (NP) and Verb Phrase (VP) are used in keyword extraction.

### B. Current tools in NLP and POS Tagging

Currently, Stanford CoreNLP (version 3.6.0) (Manning *et al.*, 2014), Open NLP (version 1.6.0) (*Welcome to Apache OpenNLP*, 2013) and NLP4J (version 1.1.3) (*emorynlp/nlp4j: NLP tools developed by Emory University*, 2016) are the widely used machine learning based Open Source NLP tools for Java. These are the NLP tools with the highest level of accuracy. The NLP tool named ANNIE POS tagger (included with GATE, version 8.2) (Cunningham *et al.*, 2001) uses a rule-based approach in contrast to machine learning methods. The present research employs a machine learning based approach of NLP tools.

POS tagging is done using Tregex (Levy and Andrew, 2006) method and the Penn Treebank notation (Marcus *et al.*, 1994) is used to POS tag each word. In both cases, the tagger uses the unidirectional model, where the tag of the current word is predicted based on the tags of its neighbours. A dependency network is used to perform this task, and a word is considered as a node in the network which is directly influenced by its neighbours (Toutanova, Klein and Manning, 2003). POS tagging is made use of in the current study, to identify the keywords of a tweet.

*C. Lemmatization*

Lemmatization uses "the vocabulary and morphological analysis of words", normally aiming "to remove inflectional endings only and to return the base or dictionary form of a word, which is known as the lemma" (Manning, Ragahvan and Schutze, 2008, p. 32). For example, this technique is used to obtain the common root "eat" from the following list.

E.g.; The lemma of 'eats', 'eating' and 'eat' is 'eat'

Another similar technique of obtaining the root is through stemming. (Manning, Ragahvan and Schutze, 2008, p. 32). In this research, the Lemma is used to expand and isolate the subject and the verb in a subject-verb contraction. Stanford CoreNLP Suite (*Stanford Named Entity Recognizer (NER)*, no date) comes with the Lemma bundled which is used for this study.

*D. Named Entity Recognition (NER) and NLP*

NER is used to extract relevant information from a text and sort them into classes. The NER (Finkel, Grenager and Manning, 2005) included in the Stanford CoreNLP suite has the ability to label words into a 7-class model. The 7 classes consist of location, person, organization, money, percent, date and time (*Stanford Named Entity Recognizer (NER)*, no date). The NER utilizes the POS tag and lemma of a word to assign a class into it. The class of a word is also used in this study to identify keywords.

*E. Use of NLP in Twitter*

As tweets are limited to 140 characters, they tend to "exhibit much more language variation. (Bontcheva *et al.*, 2013). This is one reason why previous researchers state that tweets cannot be analysed using basic POS tagging (Bontcheva *et al.*, 2013). Several attempts such TwitIE (Bontcheva *et al.*, 2013) and TweetNLP (Owoputi *et al.*, 2013) have been made to develop models to analyse POS tags of Tweets, but with limited success. In the Twitter-POS tagger model released for Stanford CoreNLP (Derczynski, Ritter, *et al.*, 2013), some of the aforementioned functionalities have been incorporated. However, there is a need for developing accurately modelled domain specific corpora, for the analysis of POS tagging in Tweets.

In the present study, "rule-based grammars for the syntactic-semantic analysis of word forms and sentences" (Hausser, 2014) is applied to extract the relevant keywords from the tweets.

*F. Turing Test and NLP*

The Turing Test (Turing, 1950) introduced by Alan Turing in 1950 is conducted to answer the question, "Can Machines Think?" (Copeland, 2004, p. 479). Thus, according to Turing, "any machine that plays the imitation game successfully can appropriately be described as a brain" (Copeland, 2004, p. 479). The participants of the test are a human respondent, a human evaluator and the machine. The test is conducted by asking a question from the human respondent and the machine, then the human evaluator is asked to identify the machine generated response out of the responses from the human and the machine (Turing, 1950; Witten, Bell and Fellows, 1998). If the human is unable to identify more than half of the machine generated responses, the machine passes the Turing Test (Witten, Bell and Fellows, 1998). Since the objective of this research is to enable the machine to imitate a human, the Turing Test was used as the method of evaluation.

*G. The Present Research*

This research was developed in two stages. The research is the extension of Stage 1 of the system.

*1) Stage 1:* This stage extracted the keywords by considering the noun phrase (NP) and verb phrase (VP). The keywords were then sent through a parser to remove any linguistic and domain specific noise, followed by another parser to include any domain specific words that were not extracted from the tweet. The method was evaluated using the Turing Test which consisted of a sample of 6 pairs (Weerasooriya, Perera and S.R. Liyanage, 2016).

*2) Stage 2:* The second stage addressed the shortcomings of stage one. The improvements needed were identified by comparing the machine generated responses and the human generated responses of stage 1.

III. METHODOLOGY

The process of developing the stages 1 and 2 is mentioned below.

*A. Data Collection and Development of the Corpora*

The Dialog Axiata (*Dialog Axiata (@dialoglk) | Twitter*, no date) Twitter profile was used to build the system. Two corpora were built by analysing the tweets. Terms that refer to Dialog Axiata and its services are identified as Domain Specific Keywords (DSK). The corpus of DSK was manually collected by analysing tweets from the month of February, 2016 to March, 2016. The corpus contained 206 words.

The domain specific 'words to reject' corpus consisted of words that do not contribute to the meaning (E.g. hello, hi, dear), interjections that are often wrongly tagged as verbs (E.g. please, thanks), certain nouns, verbs, and auxiliary verbs. This corpus consisted of 70 words.

*B. NLP and POS Tagger*

A machine learning-based POS tagger was selected for this research, as it has the ability to "exploit labelled training data to adapt to new genres or even languages, through supervised learning" (Derczynski, Ritter, *et al.*, 2013). The highest token accuracy of 97.64% is recorded by NLP4J (Nanavati and Ghodasara, 2015; *POS Tagging (State of the art)*, 2016). However, this accuracy is at stake in Twitter analysis. As a result, the token accuracy of the POS tagger declines from 97-98% to 70-75% (Derczynski, Maynard, *et al.*, 2013).

A POS Tagger model specifically trained for tweets displayed a token accuracy to 90.5% (Derczynski, Ritter, *et al.*, 2013). Out of the above mentioned list of tools, this model was available only for Stanford CoreNLP (Derczynski, Ritter, *et al.*, 2013). Hence, Stanford CoreNLP was used for the present research.

The flow chart of the methodology for Stage 1 which was developed in 2016 (Weerasooriya, Perera and S.R. Liyanage, 2016) is shown in Fig 1.

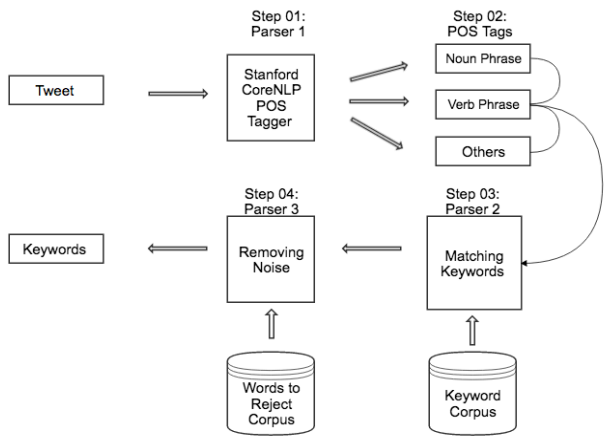

Fig. 1 Flow Chart of Stage 1

### C. Stage 2: System Design

The result of Stage 1 for an example tweet is given below.

Tweet - @dialoglk I made my payment just after my line got barred in themorning! And still the line hasn't got connected, Whats with the delay?

**Keywords – made (VBD), payment(NN), line(NN), got(VBD), barred(VBD), morning(NN), line(NN), got(VBD), connected(VBN), delay(NN)**

(Abbreviations of the Penn Treebank Notation (Marcus *et al.*, 1994):
CC – Coordination Conjunction, CD – Cardinal Number, DT – Determiner, IN – Preposition or subordination, JJ – Adjective, NN – Noun, Singular or Mass, VB – Verb, base form, VBD – Verb, past tense, VBN – Verb, past participle, VBZ – Verb, 3$^{rd}$ person singular present, PRP - Personal Pronoun, PRP$ - Possessive Pronoun, RB – Adverb, USR – Username, WP – Wh-pronoun)

The lapses identified from Stage 1 are as follows:
  i. Unnecessary time indicators included– The word "morning"
  ii. Contractions not expanded– Contractions such as "ve" and "n't"
  iii. Negation markers absent– The word "has<u>n't</u>"
  iv. Duplicate Keywords not removed– Repetition of the word "got"

The above issues were addressed using the following methods:
  i. NER - To identify and remove time indicators.
  ii. Lemma – To expand and analyse the contractions.
  iii. Adverbs – To include negation markers
  iv. LinkedHashSet – To remove duplicates

Stage 2 was also evaluated using the Turing Test and the results were recorded. Flow chart of the improved system is shown in Fig. 2.

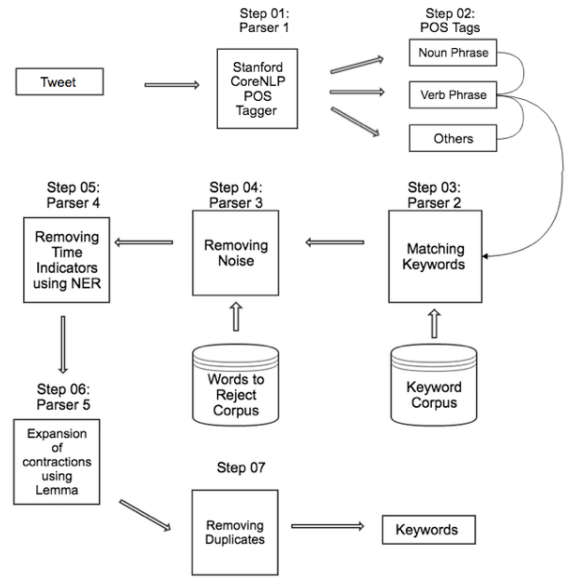

Fig. 2 The Flow Chart of the Improved System

### D. Parser 4 - NER and Time Indicators

The time indicators are, in most cases, not essential for the meaning and were not present in the human generated keyword sets. Thus, using NER, the time indicators such as "morning" were removed in the system revision. The result of the Parser 4 is shown in Fig. 3.

> Tweet - @dialoglk I made my payment just after my line got barred in the morning! And still the line hasn't got connected, Whats with the delay?
>
> **Verbs - made(VBD), got(VBD), barred(VBN), ~~has(VBZ)~~, got(VBD), connected(VBN)**
> **Nouns - payment(NN), line(NN), ~~morning(NN)~~, line(NN), delay(NN)**
> ~~Other - @dialoglk(USR), I(PRP), my(PRP$), just(RB), after(IN), my(PRP$), in(IN), the(DT), and(CC), still(RB), the(DT), n't(RB), whats(WP), with(IN), the(DT)~~

Fig. 3 The result from Parser 4

### E. Parser 5 – Expansion and Analysis of Contractions

As lemma gives the common root of a word, it was used to expand and analyse contractions. In the example (Fig. 3), the contraction "hasn't" is expanded to "has" (VBZ) and "not" (RB) through lemmatization. However, "has" is rejected as it is an auxiliary verb (see Parser 3). The result from Parser 5 is shown in Fig. 4.

```
Tweet - @dialoglk I made my payment just after my line got barred in the
morning! And still the line hasn't got connected, Whats with the delay?

Verbs - made(VBD), got(VBD), barred(VBN), ~~has(VBZ)~~, got(VBD),
connected(VBN)
Nouns - payment(NN), line(NN), ~~morning(NN)~~, line(NN), delay(NN)
~~Other - @dialoglk(USR), I(PRP), my(PRP$), just(RB), after(IN),
my(PRP$), in(IN), the(DT), and(CC), still(RB), the(DT),~~ **not(RB)**,
~~whats(WP), with(IN), the(DT)~~
```

Fig. 4 The result from Parser 5

*F. Adjustment of Parser 1 – Negation Markers (Adverbs)*

The negation markers such as 'not' from Parser 5 are identified as adverbs. When the lemma of the contractions from Parser 5 separated the 'not' from the rest of the verb, the word 'not' is included back into the keywords list as it is important for the meaning.

*G. Removing Duplicates*

The keyword list shown in Fig. 4 consists of duplicates such as "got". This was removed from the final keywords list by using a LinkedHashList, which does not allow duplicates, while retaining the sequential order in the list.

```
Tweet - @dialoglk I made my payment just after my line got barred in the
morning! And still the line hasn't got connected, Whats with the delay?

Verbs - **made(VBD), got(VBD), barred(VBN)**, ~~has(VBZ)~~, **got(VBD)**,
**connected(VBN)**
Nouns - **payment(NN), line(NN)**, ~~morning(NN)~~, **line(NN), delay(NN)**
Adverbs - **not(RB)**
~~Other - @dialoglk(USR), I(PRP), my(PRP$), just(RB), after(IN),
my(PRP$), in(IN), the(DT), and(CC), still(RB), the(DT),whats(WP),
the(DT)~~
```

Fig. 5 The final resulting keyword list

**Keywords – made (VBD), payment(NN), line(NN), got(VBD), barred(VBD), line(NN), not(RB), connected(VBN), delay(NN)**

## IV. RESULTS AND DISCUSSION

The system (also referred as the 'machine' in this section) was evaluated with and without improvements. The system without modifications is the Stanford CoreNLP (Manning *et al.*, 2014) with the Twitter model (Derczynski, Ritter, *et al.*, 2013). The system with modifications is the system presented in this research and it is tested against the Stanford CoreNLP with the Twitter model. The three systems are referred to as follows in the rest of the paper.

System A: Stanford CoreNLP with the Twitter model
System B: Stage 1 of the machine
System C: Stage 2 of the machine, KeyXtract

The systems were evaluated using two methods as follows,
a) Turing Test – To evaluate if the machine could successfully imitate human logic.
b) Performance Test – To evaluate the performance of keyword extraction by the machine. This was measured using the $F_1$ Score (Derczynski, 2013).

The Turing Test involves asking a set of questions from a human and the machine. Their answers are then evaluated by a human supervisor. If the supervisor is unable to identify the machine in at least half of the test cases, the machine passes the Turing Test (Turing, 1950). The testing was conducted in three phases,
a) To evaluate System A.
b) To evaluate System B.
c) To evaluate System C.

The performance was measured by comparing the keywords sets generated by the machine with two keywords sets produced by an English language expert and an author of the research. The human generated keyword sets were compared with the System A and System C using the $F_1$ Scores. This was used to measure the performance of the machine.

*A. Turing Test: Evaluation Methodology: Design*

The evaluations based on the Turing Test were conducted for each system as mentioned above. The System A and System B were tested with the same set of supervisors with a time gap of 3 months between the tests. This time gap was to ensure the responses would not be fresh in the minds of the human participants. Since the System C was improved considering the previous responses, it was tested with a new set of supervisors.

*B. Turing Test: Evaluation Methodology: Participants*

All three systems were tested with six test cases (each consisting of the machine, the human keyword generator and a human supervisor). The human participants in the six test cases were chosen to represent six different fields. The criteria of the test cases are given in Table I.

Keywords sets were produced by the human and the machine, and the human supervisor tried to identify the system generated answers.

TABLE I
TEST CASE CRITERIA

| Test Case Number | Test Case Criterion | Justification | Minimum Requirement |
|---|---|---|---|
| 1 | Academics | Frequent users of Academic English | University lecturers who are not from the field of English |
| 2 | English Language Experts | Competent in English language and literature | English Language Lecturers |
| 3 | Undergraduates | Use English for academic purposes | Individuals currently reading for a Bachelor's degree |
| 4 | Graduates | Use English in a professional context | Individuals who have completed a Bachelor's Degree |
| 5 | Computer Science Graduates | Have an expert knowledge in Computer Science | Computer Science graduates working in the industry |

| Test Case Number | Test Case Criterion | Justification | Minimum Requirement |
|---|---|---|---|
| 6 | Randomly selected twitter users | Being Familiar with Twitter | Twitter Users |

*C. Turing Test: Evaluation Methodology: Keyword Extraction*

The testing dataset consisted of 14 tweets (2 tweets per day collected for 7 days). They were collected from the first week of April, 2016 (3rd April to 9th April). The dataset contained a new set of tweets. The set of tweets were given to the human keyword generators to extract keywords and then to the machine to do the same. This was repeated for Systems A, B and C.

*D. Turing Test: Evaluation Methodology: Evaluation of Test Cases*

The responses generated at the extraction phase by the human keyword generators, the Systems A, B and C were used in this stage.

In the first round, the supervisor was provided with the original tweet and the two sets of keywords generated by the System A and the human. The supervisor was asked to identify the set of keywords which was generated by the System A (the machine).

The same process was repeated with the keywords sets generated by the human and System B, and the human and the System C. System C was tested last with a group of fresh supervisors who were completely new to the research.

*E. Turing Test: Evaluation Discussion: Identical Keyword Extraction*

During the keyword extraction phase, all three systems produced several keyword sets which were identical to the responses of the human.

An example for this occurrence is given below,

"@dialoglk Where i can buy a touch travel pass?"

Machine Generated Keywords - buy, touch, travel, pass
Human Generated Keywords - buy, touch, travel, pass
Among System B, and C, 5 tweets out of 14 had keyword sets where the answers of the human and the System were identical.

*F. Turing Test: Evaluation Discussion: Results*

A summary of the overall evaluation results is given in Tables III (for System A), Table IV (for System B) and Table V (for System C). The total instances where the machine was successful was calculated using the given formula.

$$T = \left(\frac{x + z}{n}\right) * 100\%$$

where,
T - Total instances where the system was successful
x – Instances where the Machine and Human answers are identical
z – Instances where the Supervisor did not detect the answer generated by the Machine
n - Total number of tweets

Evaluation Results of System C is shown in Table II.

TABLE II
SUMMARY OF TURING TEST APPLIED FOR SYSTEM C: STAGE 2 OF THE MACHINE

| Test Case Criterion | Machine and Human answers were identical (x) | Supervisor detected the answer generated by the Machine (y) | Supervisor could not detect the answer generated by the machine (z) | Total instances where the system was successful (T) |
|---|---|---|---|---|
| Academics | 0 | 11 | 3 | 21.43% |
| English Language Experts | 0 | 4 | 10 | 71.43% |
| Undergraduates | 4 | 8 | 2 | 85.71% |
| Graduates | 4 | 7 | 3 | 50.00% |
| Computer Science Graduates | 4 | 3 | 7 | 78.57% |
| Randomly selected twitter users | 5 | 5 | 4 | 64.29% |

The machine was unsuccessful only with academics, this could be due to their familiarity with analytical and technical writing.

An overview of the Turing test results is shown in Table III.

TABLE III
COMPARISON OF THE TURING TEST RESULTS

| System Tested | Test cases that passed | Test cases that failed | Success rate of the System |
|---|---|---|---|
| System A: Stanford CoreNLP with the Twitter model | 3 | 3 | 50.00% |
| System B: Stage 1 of the machine | 5 | 1 | 83.33% |
| System C: Stage 2 of the machine | 5 | 1 | 83.33% |

According to the Table VI, it is evident that the modified systems have more success in imitating the human in extracting keywords than the system without any modifications.

*G. Performance Test: $F_1$ Score Evaluation Discussion*

The performance of the machine was evaluated using the $F_1$ Score. Initially, an English Language expert (ELE) and an author of the paper generated the controller dataset of keywords from the 14 tweets used for the Turing Test, from Section C. Two human generated keyword sets were used factoring the subjectivity of the keyword extraction process. The average of the $F_1$ Scores from the two sets of keywords was used for the evaluation.

The $F_1$ Score(F) was calculated by analysing the keywords generated for each tweet according to the formula (Derczynski, 2013) given below,

$$F_1 = 2 * \left(\frac{P * R}{P + R}\right)$$

where,
F – $F_1$ Score
P - Precision
R – Recall

The precision (P) was computed by dividing the true positives (i.e. the number words which were common to the human and the machine data set) by the false positives (i.e. the total number words which were extracted by the machine). The recall (R) was computed by dividing the true positives by the total number of words which were extracted by the human.

The $F_1$ score was computed for System A and System C. The results for dataset by the ELE is included in Table IV.

TABLE IV
$F_1$ SCORES FOR ENGLISH LANGUAGE EXPERT'S DATASET

| Tweet# | Word# | System A | | | System C | | |
|---|---|---|---|---|---|---|---|
| | | P | R | $F_1$ | P | R | $F_1$ |
| 1 | 9 | 0.40 | 1.00 | 0.57 | 0.50 | 1.00 | 0.67 |
| 2 | 11 | 0.43 | 0.75 | 0.55 | 0.60 | 0.75 | 0.67 |
| 3 | 25 | 0.38 | 0.60 | 0.46 | 0.50 | 0.80 | 0.62 |
| 4 | 15 | 0.71 | 1.00 | 0.83 | 0.80 | 0.80 | 0.80 |
| 5 | 24 | 0.55 | 0.86 | 0.67 | 0.44 | 0.57 | 0.50 |
| 6 | 25 | 0.23 | 0.75 | 0.35 | 0.36 | 1.00 | 0.53 |
| 7 | 25 | 0.25 | 0.40 | 0.31 | 0.45 | 1.00 | 0.63 |
| 8 | 23 | 0.25 | 0.67 | 0.36 | 0.60 | 1.00 | 0.75 |
| 9 | 9 | 1.00 | 1.00 | 1.00 | 1.00 | 1.00 | 1.00 |
| 10 | 7 | 1.00 | 1.00 | 1.00 | 1.00 | 1.00 | 1.00 |
| 11 | 7 | 1.00 | 1.00 | 1.00 | 1.00 | 1.00 | 1.00 |
| 12 | 20 | 0.30 | 1.00 | 0.46 | 0.43 | 1.00 | 0.60 |
| 13 | 13 | 0.83 | 1.00 | 0.91 | 0.83 | 1.00 | 0.91 |
| 14 | 10 | 1.00 | 0.71 | 0.83 | 1.00 | 0.86 | 0.92 |
| Average | | 0.59 | 0.84 | **0.66** | 0.68 | 0.91 | **0.76** |

The results show that the System C has improved from a $F_1$ score of 0.66 to 0.76.

The $F_1$ score computed for the dataset extracted by an author of the paper is included in Table V below,

TABLE V
$F_1$ SCORES FOR THE DATASET BY AN AUTHOR

| Tweet# | Word# | System A | | | System C | | |
|---|---|---|---|---|---|---|---|
| | | P | R | $F_1$ | P | R | $F_1$ |
| 1 | 9 | 0.47 | 1.00 | 0.57 | 0.50 | 1.00 | 0.67 |
| 2 | 11 | 0.57 | 1.00 | 0.73 | 0.80 | 1.00 | 0.89 |
| 3 | 25 | 0.38 | 0.60 | 0.46 | 0.50 | 0.80 | 0.62 |
| 4 | 15 | 0.71 | 1.00 | 0.83 | 0.80 | 0.80 | 0.80 |
| 5 | 24 | 0.55 | 0.86 | 0.67 | 0.44 | 0.67 | 0.53 |
| 6 | 25 | 0.23 | 0.75 | 0.35 | 0.36 | 1.00 | 0.53 |
| 7 | 25 | 0.50 | 0.57 | 0.53 | 0.45 | 0.71 | 0.56 |
| 8 | 23 | 0.38 | 0.75 | 0.50 | 0.80 | 1.00 | 0.89 |
| 9 | 9 | 1.00 | 0.80 | 0.89 | 1.00 | 0.80 | 0.89 |
| 10 | 7 | 1.00 | 1.00 | 1.00 | 1.00 | 1.00 | 1.00 |
| 11 | 7 | 1.00 | 1.00 | 1.00 | 1.00 | 1.00 | 1.00 |
| 12 | 20 | 0.30 | 1.00 | 0.46 | 0.43 | 1.00 | 0.60 |
| 13 | 13 | 0.83 | 1.00 | 0.91 | 1.00 | 1.00 | 1.00 |
| 14 | 10 | 1.00 | 0.83 | 0.91 | 1.00 | 1.00 | 1.00 |
| Average | | 0.63 | 0.87 | **0.70** | 0.72 | 0.91 | **0.78** |

The $F_1$ score of the modified System C has increased from 0.70 to 0.78. The average of the two $F_1$ Scores obtained from the two sets are summarized in the Table VI,

TABLE VI
SUMMARY OF $F_1$ SCORES

| System Tested | ELE Dataset | Author Dataset | Average |
|---|---|---|---|
| System A: Stanford CoreNLP with the Twitter model | 0.66 | 0.70 | 0.69 |
| System C: Stage 2 of the machine | 0.76 | 0.78 | 0.77 |

The highest $F_1$ scores are recorded from the Tweets which of shorter length, proving that the accuracy of the machine is high in short tweets.

V. CONCLUSION

The traditional machine learning based NLP tools have failed to accurately classify the tokens of the tweet with POS tags.

The paper presents the Twitter Model of KeyXtract, a system developed with a mix of machine learning and a rule-based approaches. A combination of the Stanford CoreNLP and the Twitter-POS tagger model was used for POS tagging to extract keywords from the tweet. The rule based approach was used to remove unnecessary words from the initial word group selection with the help of corpora. The research was developed in two stages. Stage 2 included modifications to Stage 1 such as using NER to remove time indicators and measures to include negation markers.

The research was evaluated using two methods. The ability to imitate the human logic in extracting keywords was measured with the help of the Turing Test while the performance was measured using the $F_1$ Score.

The final modified system passed the Turing Test with an overall result of 83.33%There were more instances where the modified system produced the same set of results as humans. Since the system from stage 3 consists of the improvements made to the system from stage 2, the evaluation results look quite promising. The system could be tested with a larger population for nuance results.

The performance measures of the system showed that the $F_1$ scores increased from 0.69 of system A (system without any modifications) to 0.77 of system C (final system with modification). It was also evident that the system's level of precision was high in analysing short tweets.

Future work in the research could include the use of a complete domain specific corpus and the ability to analyse emoji, which would improve the accuracy of the keywords extracted by keyword matching. As this approach uses linguistic features to extract keywords, the same approach could be applied to other NLP tools

ACKNOWLEDGMENT

The authors would like to thank the participants of the Turing Test for their time and effort to evaluate KeyXtract.

REFERENCES

Bontcheva, K., Derczynski, L., Funk, A., Greenwood, M. A., Maynard, D. and Aswani, N. (2013) 'TwitIE: An Open-Source Information Extraction Pipeline for Microblog Text'.

Copeland, B. J. (2004) *The Essential Turing: Seminal Writings in Computing, Logic, Philosophy, Artificial Intelligence, and Artificial Life: Plus The Secrets of Enigma*. Edited by B. J. Copeland. Oxford: Clarendon Press.

Cunningham, H., Maynard, D., Bontcheva, K. and Tablan, V. (2001) 'Gate', *Proceedings of the 40th Annual Meeting on Association for Computational Linguistics - ACL '02*, (July), p. 168. doi: 10.3115/1073083.1073112.

Derczynski, L. (2013) 'Complementarity, F-score, and NLP Evaluation'.

Derczynski, L., Maynard, D., Aswani, N. and Bontcheva, K. (2013) 'Microblog-genre noise and impact on semantic annotation accuracy', *Ht 2013*, (May), pp. 21–30. doi: 10.1145/2481492.2481495.

Derczynski, L., Ritter, A., Clark, S. and Bontcheva, K. (2013) 'Twitter part-of-speech tagging for all: Overcoming sparse and noisy data', *Proceedings of the Recent Advances in Natural Language Processing*, (September), pp. 198–206. Available at: http://www.aclweb.org/website/old_anthology/R/R13/R13-1026.pdf.

*Dialog Axiata (@dialoglk) | Twitter* (no date). Available at: https://twitter.com/dialoglk.

*emorynlp/nlp4j: NLP tools developed by Emory University* (2016). Available at: https://github.com/emorynlp/nlp4j.

Finkel, J. R., Grenager, T. and Manning, C. (2005) 'Incorporating non-local information into information extraction systems by gibbs sampling', *in Acl*, (1995), pp. 363–370. doi: 10.3115/1219840.1219885.

Hausser, R. (2014) *Foundations of Computational Linguistics*. Third Edit. London: Springer Heidelberg. doi: 10.1007/978-3-662-04337-0.

Levy, R. and Andrew, G. (2006) 'Tregex and Tsurgeon: tools for querying and manipulating tree data structures', *5th International Conference on Language Resources and Evaluation (LREC 2006)*, pp. 2231–2234.

Manning, C. D., Bauer, J., Finkel, J., Bethard, S. J., Surdeanu, M. and McClosky, D. (2014) 'The Stanford CoreNLP Natural Language Processing Toolkit', *Proceedings of 52nd Annual Meeting of the Association for Computational Linguistics: System Demonstrations*, pp. 55–60. Available at: http://aclweb.org/anthology/P14-5010.

Manning, C. D., Ragahvan, P. and Schutze, H. (2008) *An Introduction to Information Retrieval*. 1st edn. New York, NY: Cambridge University Press. doi: 10.1109/LPT.2009.2020494.

Marcus, M., Kim, G., Marcinkiewicz, M. A., MacIntyre, R., Bies, A., Ferguson, M., Katz, K. and Schasberger, B. (1994) 'The Penn Treebank: Annotation Predicate Argument Structure', *Proceedings of the workshop on Human Language Technology - HLT '94*, pp. 114–119. doi: 10.3115/1075812.1075835.

Mitkov, R. and Ha, L. A. (1999) 'Computer-Aided Generation of Multiple-Choice Tests', *Proceedings of the HLT-NAACL 03 workshop on Building educational applications using natural language processing*, pp. 17–22.

Nanavati, J. and Ghodasara, Y. (2015) 'A Comparative Study of Stanford NLP and Apache Open NLP in the view of POS Tagging', *International Journal of Soft Computing and Engineering (IJSCE)*, 5(5), pp. 57–60. Available at: http://www.ijsce.org/attachments/File/v5i5/E2744115515.pdf.

Owoputi, O., O 'connor, B., Dyer, C., Gimpel, K., Schneider, N. and Smith, N. A. (2013) 'Improved Part-of-Speech Tagging for Online Conversational Text with Word Clusters', *Proceedings of NAACL*.

*POS Tagging (State of the art)* (2016). Available at: http://aclweb.org/aclwiki/index.php?title=POS_Tagging_(State_of_the_art) (Accessed: 22 August 2016).

*Stanford Named Entity Recognizer (NER)* (no date). Available at: http://nlp.stanford.edu/software/CRF-NER.html (Accessed: 20 November 2016).

Toutanova, K., Klein, D. and Manning, C. D. (2003) 'Feature-rich part-of-speech tagging with a cyclic dependency network', *In Proceedings of the 2003 Conference of the North American Chapter of the Association for Computational Linguistics on Human Language Technology - Volume 1 (NAACL '03)*, pp. 252–259. doi: 10.3115/1073445.1073478.

Turing, A. M. (1950) 'Computing Machinery and Intelligence', *Mind*, 49, pp. 433–460. doi: 10.1093/mind/LIX.236.433.

*Twitter Usage / Company Facts* (2016). Available at: https://about.twitter.com/company (Accessed: 30 April 2017).

Weerasooriya, T., Perera, N. and Liyanage, S. R. (2016) 'A method to extract essential keywords from a tweet using NLP tools', in *2016 Sixteenth International Conference on Advances in ICT for Emerging Regions (ICTer)*. IEEE, pp. 29–34. doi: 10.1109/ICTER.2016.7829895.

Weerasooriya, T., Perera, N. and Liyanage, S. R. (2016) 'A Method to Extract Essential Keywords from a Tweet using NLP Tools', in *2016 International Conference on Advances in ICT for Emerging Regions (ICTer)*. Colombo, pp. 29–34.

*Welcome to Apache OpenNLP* (2013). Available at: http://opennlp.apache.org/.

Witten, I. H., Bell, T. and Fellows, M. (1998) *Computer Science Unplugged . . .Off-Line Activities and Games for All Ages*.

Zitouni, I. (ed.) (2014) *Natural Language Processing of Semitic Languages*. New York, NY: Springer Berlin Heidelberg. doi: 10.1007/978-3-642-45358-8.